\documentclass[10pt, conference, compsocconf]{IEEEtran}
\usepackage{color,soul}
\usepackage{cite}
\ifCLASSINFOpdf
  \usepackage[pdftex]{graphicx}
\else
\fi

\begin{document}
%
\title{Deep Weakly Supervised Domain Adaptation for \\ Pain Localization in Videos}


\author{\IEEEauthorblockN{Gnana Praveen R, Eric Granger, Patrick Cardinal}
\IEEEauthorblockA{Laboratory of Imaging, Vision and Artificial Intelligence (LIVIA)\\
Dept. of Systems Engineering, École de technologie supérieure\\
Montreal, Canada\\
}
}


%

\maketitle
\begin{abstract}
Automatic pain assessment has an important potential diagnostic value for populations that are incapable of articulating their pain experiences. 
As one of the dominating nonverbal channels for eliciting pain expression events, facial expressions has been widely investigated for estimating the pain intensity of individual. However, using state-of-the-art deep learning (DL) models in real-world pain estimation applications poses several challenges related to the subjective variations of facial expressions, operational capture conditions, and lack of representative training videos with labels. 
Given the cost of annotating intensity levels for every video frame, we propose a  weakly-supervised domain adaptation (WSDA) technique that allows for training 3D CNNs for spatio-temporal pain intensity estimation using weakly labeled videos, where labels are provided on a periodic basis. 
In particular, WSDA integrates multiple instance learning into an adversarial deep domain adaptation framework to train an Inflated 3D-CNN (I3D) model such that it can accurately estimate pain intensities in the target operational domain. The training process relies on weak target loss, along with domain loss and source loss for domain adaptation of the I3D model. Experimental results obtained using labeled source domain RECOLA videos and weakly-labeled target domain UNBC-McMaster videos indicate that the proposed deep WSDA approach can achieve significantly higher level of sequence (bag)-level and frame (instance)-level pain localization accuracy than related state-of-the-art approaches. 

\end{abstract}

%
\IEEEpeerreviewmaketitle

\section{Introduction}
Pain is a vital indicator of our health condition that places an enormous economic burden on health care systems. 
It is usually self-reported by patients, either through clinical inspection, or using the Visual Analog Scale (VAS). However, self-reported pain assessment is vulnerable to bias, and cannot be used for people incapable of articulating their pain experiences. Automatic pain assessment has an important potential diagnostic value for people, such as infants, young children and persons with communicative or neurological impairments.
At present, health professionals must infer pain in individuals by examining various physiological and behavioural indicators that are strongly associated with pain.

Human face is a rich source for non-verbal information regarding a person's health, and facial expression can be considered as a reflective and spontaneous reaction of painful experiences. The estimation of pain can be formulated as classification or regression problem. Most of the early approaches \cite{sikka, chongliang} addressed the problem of pain estimation as a classification problem, where binary pain/no-pain or discrete levels are detected. However, regression based approaches have recently gained attention as they provide wide range of continuous intensity values i.e., valence and arousal, which is crucial for pain localization in videos.  
This paper focuses on deep learning (DL) models for pain intensity estimation from faces captured in videos.  
Recently, dynamic or spatio-temporal FER techniques have emerged as a promising approach to improve performance, where expression is estimated from a sequence of consecutive frames in a video \cite{Pan06}.

Deep Learning (DL) models in particular CNNs, provide state-of-the-art performance using large amount of labeled training data in many visual recognition applications such as image classification, object detection and segmentation. Several models have been developed to aggregate the network output produced using 2D-CNN for consecutive facial regions of interest (ROIs) in a video, either through decision-level or feature-level frame aggregation. However, 3D CNNs are found to be efficient in capturing the spatiotemporal dynamics \cite{Mel19}. 
Some important challenges remain for the design of accurate DL models for real-world pain estimation applications. 
Pain expressions are sparse in nature as they occur less frequently, resulting in limited data-sets that can exploit the potential of DL models. Collecting and annotating large-scale data sets that contain controlled expressions is a costly and time-consuming process requiring domain experts. Moreover, manual annotation according to the intensity of an expression, or to the levels of valence and arousal, is a complex process that is subjective in nature, resulting in ambiguous annotations.
In practice, it is not feasible to annotate each sample of a continuous audio-visual signals, e.g., each video frame. Therefore, techniques for weakly-supervised learning (WSL) are gaining attention to handle the problem of data with limited annotations. 

Given the cost and challenges of annotating data, techniques for weakly-supervised learning (WSL) are very appealing 
as they allow exploiting weak labels to train DL models. They can be applied in scenarios involving incomplete supervision, inexact supervision, and ambiguous or inaccurate supervision \cite{Zhou2017}. The inexact supervision scenario is relevant to our application, where training data-sets only require global annotations for an entire video, or on a periodic basis for video sequences. 
Multiple Instance Learning (MIL) is one of the widely used approaches for inexact supervision \cite{MILSurvey}. However, existing MIL based approaches for automatic pain intensity estimation are based on traditional machine learning approaches due to the lack of sufficient data as facial expressions pertinent to pain are sparse in nature. Though MIL is used for both bag level and instance level prediction, we have primarily focused on instance-level prediction for pain localization in videos. 

Another key challenge to develop DL models is significant variations of facial expressions over different persons and operational capture conditions. In real-world applications, performance of DL models decline if there is considerable divergence or shift between the capture conditions in development and operational environments \cite{Lia18}. For accurate pain estimation, DL models must therefore be adapted to a specific target domain – person and capture condition (sensors, computing device, and environment). 
DA techniques have been introduced to diminish the impact on performance caused by domain shifts between source and target data distributions and the problem of limited dataset by learning discriminant and domain-invariant representations. 

In this paper, we propose a deep weakly-supervised domain adaptation (WSDA) technique to train 3D-CNN models for regression to address the challenges of pain intensity estimation from videos with limited annotations. Deep Adversarial DA is performed with weak sequence-level labels. It is assumed that the source domain dataset is comprised of full-annotated videos collected during development, and target domain videos are annotated periodically (at the sequence level) with pain intensity values. In particular, WSDA performs multiple instance learning within a deep DA framework, where inflated 3D-CNN (I3D) model is trained using weak target loss along with domain and source losses for accurate estimation of pain intensities in the target operational domain.

\section{Related Work}
\subsection{Deep Learning Models for Pain Localization:}
Most DL models proposed for pain intensity estimation has been explored in the context of fully supervised learning. Wang et al \cite{8296449} addressed the problem of limited data-set of facial expressions by fine-tuning the face recognition network using a regularized regression loss. 
Rodrigue et al \cite{7849133} used VGG Face pre-trained CNN network \cite{Parkhi15} for capturing the facial features and further fed to the LSTM network to exploit the temporal relation between the frames. Apart from 2D models with LSTM, researchers also explored optical flow and 3D CNN based approaches for temporal modeling of facial expressions. 
Jing et al \cite{ZhouHSZ16} proposed Recurrent Convolutional Neural Network (RCNN) by adding recurrent connections to the convolutional layers of the CNN architecture for estimating the pain intensities. However, choosing fixed temporal kernel depth fails to capture varying levels of temporal ranges as duration of facial expressions may vary from short to long temporal ranges. In order to address the problem of fixed temporal depth, Tavakolian et al \cite{Tavakolian2018DeepSR} designed a novel 3D CNN based architecture using a stack of convolutional modules with varying kernel depths for efficient dynamic spatiotemporal representation of faces in videos. 

Compared with 2D models, 3D-CNNs are found to be quite promising in capturing the temporal dynamics of the video sequences. The limitation of these approaches is that they require frame level intensity labels, which is a major bottleneck in real-time scenarios. To overcome this problem, we have used fully supervised source domain to compensate the problem of limited annotations with deep learning models using adversarial domain adaptation. Inspired by the performance of 3D CNN models, we have used 3D CNN model i.e., I3D \cite{8099985} in particular for modeling spatiotemporal dynamics of facial expressions.

\subsection{Deep Domain Adaptation:}
Unsupervised DA has been widely used for many applications related to facial analysis such as face recognition, facial expression recognition, smile detection, etc. Their accuracy significantly degrades when there is a domain shift in the test images which can be due to variations in pose, illumination, resolution, blur, expressions and modality. 
Wang et al \cite{Xiaoqing} proposed unsupervised domain adaptation approach for small target data-set using Generative Adversarial Networks (GAN), where GAN generated samples are used to fine-tune the model pretrained on the source data-set. 
Contrary to the existing DA approaches for facial expression analysis, we have explored domain adaptation in the context of adapting source data with full labels to target data with coarse labels. Ganin et al \cite{Ganin:2015:UDA:3045118.3045244} proposed a novel approach of adversarial domain adaptation using deep models with partial or no target data labels using a simple gradient reversal layer. We have further extended their approach for the scenario of coarsely labeled target data for pain localization in videos. 

\subsection{Multiple Instance Learning:}


Despite the recent popularity of MIL for various computer vision applications, relatively few techniques have been developed for dynamic FER. Tax et al \cite{tax} have showed that the presence of specific frames (concept frames) are sufficient for reliable detection of facial expressions in videos. Inspired by the idea of \cite{tax}, Sikka et al. \cite{sikka} extended the idea of "concept frames" to "concept segments" in MIL setting, where "concept segments" are sub-sequences (instances) with relevant expressions for detecting pain expression events in videos. An instance-level classifier is developed using MILBOOST \cite{multiple-instance-boosting-for-object-detection}. 
Chongliang et al \cite{chongliang} further enhanced the approach by incorporating discriminative Hidden Markov Model (HMM) based instance level classifier in conjunction with MIL framework instead of MILBOOST. 
Adria et al \cite{adria} formulated the problem as multi-concept MIL framework by modeling a set of $k$ hyper-planes to discover discriminative pain expressions termed as 'concepts' in a video sequence. Sikka et al \cite{lomo} investigated the temporal dynamics of facial expressions in videos in terms of ordinal relationship of templates of sequences 
by retaining the ordering of templates (pain expression events at various intesnity levels). 

One major drawback of the above mentioned approaches is that they focus on classification of pain events rather than regression. Zhang et al \cite{zhangbilateral} explored MIL based approach based on ordinal relevance, intensity smoothness and relevance smoothness based on the gradual evolving process of facial behavior assuming the availability of peak and valley frames. 
Ruiz et al \cite{midorf} proposed multi-instance dynamic ordinal random fields (MI-DORF) based on multiple instance learning for modeling temporal sequences of ordinal instances. Their problem formulation is closely related to our framework as it deals with multiple instance regression \cite{mir}. 
However, our approach is based on 3D CNN model whereas \cite{midorf} is based on ordinal random fields. 

Though pain intensity level estimation can be performed at bag (sequence) level as well as instance (frame) level, instance-level prediction conveys much more relevant information than bag-level prediction regarding the localization of pain expression events related to the specific duration. Despite the vital importance of instance level prediction, most of the prior MIL based approaches \cite{adria}, \cite{sikka}  have focused on bag-level prediction though a few approaches \cite{midorf} have been recently proposed for instance level prediction. Inspired by the crucial importance of instance-level prediction, we have primarily focused on instance-level prediction i.e., localization of pain expression events using sequence-level labels. 


\begin{figure*}
\centering
\includegraphics[width=0.46\linewidth]{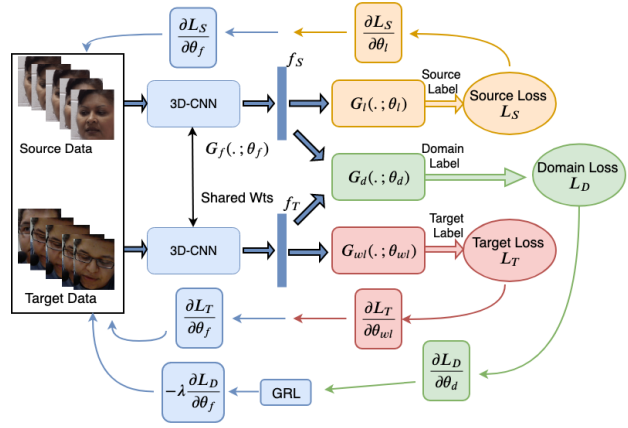}
\caption{\textbf{Overall Architecture of our proposed deep WSDA technique for training 3D CNNs on weakly-labeled target videos. Best viewed in color.}}
\label{fig:WSDDA}
\end{figure*}

\section{Deep Weakly-Supervised Domain Adaptation}

\subsection{Problem Formulation:}
Let $D = \{ ({X_1},{Y_1}),({X_2},{Y_2}),.......,({X_N},{Y_N})\} $ represents the data-set of facial expressions from videos and ${X_i}$ denotes a video sequence of the training data, which consists of set of frames as instances. Specifically, ${X_i} = \{ x_i^1,x_i^2,....x_i^{n_i}\} $ represents the temporal sequence of ${n_i}$ frames and $x_i^t$ denotes $t^{th}$ frame in $i^{th}$ sequence, where $t \in \{ 1,2,......,{n_i}\} $. ${Y_i}$ denote the labels of continuous or discrete values, which acts as frame-level and sequence-level labels for source and target domains respectively. $N$ represents the number of training samples in the data-set.  The objective of the problem is to estimate the regression model $F:X \to H$ from the training data $D$, where $X$ denotes the set of input video sequences and $H$ represents the hidden instance-level labels. The label of the sequence ${X_i}$ is predicted as structured output $H_i \in H$, where 
${H_i} = \{ h_i^1,h_i^2,.......,h_i^{n_i}\}  $ and each frame $x_i^t$ of the sequence is assigned a latent regressed value ${h_i^t}$.

In the context of Multiple Instance Regression \cite{mir}, the relationship between the bag labels $Y$ and latent ordinal states $H$ is modeled by assigning the maximum value of ${H_i}$ to the bag label ${Y_i}$ as:
\begin{equation}
{Y_i} = \mathop {\max } ({H_i}){\rm{     }}~~~~~~{H_i} = \{ h_i^1,h_i^2,.......,h_i^{n_i}\}       
\end{equation}
If the label ${Y_i}$ is $0$, then all the frames in the sequence ${X_i}$ will be assigned $0$ i.e., neutral frame. 

\subsection{Proposed Approach:}

Inspired by the idea of \cite{8099985}, we have used I3D model in our approach to incorporate the temporal information of the videos by inflating all the filters and pooling kernels of 2D architectures. In contrast with existing approaches, we have explored domain adaptation in the context of MIL, where coarse high level labels are provided instead of partial accurate labels. In the proposed framework, the deep features are learned by jointly optimizing the prediction loss and domain in-variance using adversarial domain adaptation, which minimizes the discrepancy between source and target domains still exploiting the weak labels of target data.

Let $S$ represents the source data-set, which is fully labeled data-set (frame-level labels) and $T$ denotes the target data-set, which is weakly labeled data-set (sequence-level labels). The objective is to exploit the full labels of source data and adapt the disriminative I3D model on the source data-set to the weakly labeled limited target data-set in order to predict labels of test data of the target domain. In our proposed approach of WSDA, the deep network architecture can be decomposed to three major building blocks : feature mapping, label predictor and domain classifier. Let ${G_f}$  represents the feature mapping function, where the parameters of this mapping are denoted by ${\theta _f}$. Similarly, the feature vectors of source domain and target domain are mapped to the corresponding labels using ${G_l}$ and ${G_{wl}}$, whose parameters are denoted by ${\theta _l}$ and ${\theta _{wl}}$ respectively. Finally, the mapping of feature vector to the domain label is obtained by ${G_d}$ with parameters ${\theta _d}$. In the deep network, the feature mapping layers share weights between the source and target domains to ensure common feature space between source and target domains. It has been shown that the label prediction accuracy on the target domain will be same as that of the source domain by ensuring the similarity of distributions between source and target domains \cite{SHIMODAIRA2000227}. Next, adversarial mechanism is deployed between the domain discriminator ${G_d}$, which maximizes the discrimination between source and target domains and feature extractor ${G_f}$, which is learned by minimizing the domain discrepancy between source and target domains. During training, label prediction loss is minimized on the source domain by optimizing the parameters of ${G_f}$ and ${G_l}$ in order to learn the feature mapping to corresponding labels. Simultaneously, the learned features are ensured to be domain-invariant by maximizing the loss of domain classifier to minimize the discrepancy between the source and target domains 
The label prediction loss (${L_S}$) for source domain is defined by 
\begin{equation}
{L_S} = \frac{1}{{{N_s}}}\sum\limits_{\mathop {i = 1}\limits_{{d_i} = 0} }^{N_s} {  \frac{1}{{{n_i}}}  \sum\limits_{j = 1}^{{n_i}} {MSE({G_l}({G_f}(x_i^j)),y_i^j)} } 
\end{equation}
where ${{d_i} = 0}$ represents the source domain, ${N_s}$ denotes the number of sub-sequences in the source domain, ${n_i}$ denotes the number of frames in the corresponding sub-sequence and $y_i^j$ denotes the frame labels of sub-sequences. The domain classification loss to discriminate the source and target domain is a typical binary classification problem. Therefore, logistic loss function is used as domain classification loss, which is given by 
\begin{equation}
{L_d} = \frac{1}{{{N_s} + {N_T}}}\sum\limits_{\mathop {i = 1}\limits_{{d_i} = 0,1} }^{{N_s} + {N_T}} { {logistic({G_d}({G_f}(x_i)),d_i)} } 
\end{equation}

where ${N_T}$ represents the number of sub-sequences in the target domain and $d_i$ denotes the domain label of the ${i^{th}}$ sub-sequence.

Finally, the supervision of weak labels in the target domain is deployed in the feature learning mechanism, where the parameters of ${G_{wl}}$ are optimized by mapping the features vectors of target data to the corresponding weak labels still maximizing the domain discriminator loss. Since weak labels of target data are provided at sub-sequence level, the prediction loss for target domain (${L_T}$) is estimated as mean square loss between the true and predicted labels of sub-sequences. In order to align with the true label of video sequence, the predicted label of the video sequence is obtained as the maximum value of the predictions of the sub-sequences (MIR assumption), which is given by    
\begin{equation}
{L_T} = \frac{1}{{{N_T}}}\sum\limits_{\mathop {i = 1}\limits_{{d_i} = 1} }^{{N_T}} {MSE(\mathop {\max }\limits_{j = 1,..{n_i}} ({G_{wl}}({G_f}(x_i^j))),Y_i)} 
\end{equation}
where $y_i$ denotes the weak labels of the sub-sequences in target domain.


The overall loss of the deep network architecture is given by
\begin{equation}
L = {L_S} + {L_T} - \lambda {L_d}
\end{equation}
where $\lambda$ is the trade-off parameter between the objectives of label prediction loss and domain prediction loss. The parameters of ${\theta _l}$, ${\theta _{wl}}$, ${\theta _f}$ and ${\theta _d}$ are jointly optimized using Stochastic Gradient Descent (SGD) as shown below: 

\begin{equation}
{\theta _f} \leftarrow {\theta _f} - \mu (\frac{{\partial {L_S}}}{{\partial {\theta _f}}} + \frac{{\partial {L_T}}}{{\partial {\theta _f}}} - \lambda \frac{{\partial {L_d}}}{{\partial {\theta _f}}})
\end{equation}
\begin{equation}
{\theta _l} \leftarrow {\theta _l} - \mu (\frac{{\partial {L_S}}}{{\partial {\theta _l}}})    
\end{equation}
\begin{equation}
{\theta _{wl}} \leftarrow {\theta _{wl}} - \mu (\frac{{\partial {L_T}}}{{\partial {\theta _{wl}}}})
\end{equation}
\begin{equation}
{\theta _d} \leftarrow {\theta _d} - \mu (\frac{{\partial {L_D}}}{{\partial {\theta _d}}})
\end{equation}

where $\mu $ is the learning rate of the optimizer. At the end of the training, the parameters of ${\theta _l}$, ${\theta _{wl}}$, ${\theta _f}$ and ${\theta _d}$ are expected to give a saddle point for the overall loss function as given by :
\begin{equation}
{{\hat \theta }_f},{{\hat \theta }_l},{{\hat \theta }_{wl}} = \mathop {\arg \min }\limits_{{\theta _f},{\theta _l},{\theta _{wl}}} L({\theta _f},{\theta _l},{\theta _{wl}},{{\hat \theta }_d})    
\end{equation}
\begin{equation}
{{\hat \theta }_d} = \mathop {\arg \max }\limits_{{\theta _d}} L({{\hat \theta }_f},{{\hat \theta }_l},{{\hat \theta }_{wl}},{\theta _d})    
\end{equation}
At the saddle point, the feature mapping parameters ${\theta _f}$ minimize the label prediction loss to ensure discriminative features and maximizes the domain classification loss to constrain the features to be domain-invariant. For compatibility of SGD with the adversarial mechanism, a special gradient reversal layer (GRL) is used to back propagate through the negative term (-$\lambda$) in our loss function. The value of lambda is modified over successive epochs, such that the supervised prediction loss dominates at
the early epochs of training. Further details on training mechanism can be found in \cite{Ganin:2015:UDA:3045118.3045244}. The overall mechanism of our approach for weakly supervised deep domain adaptation is shown in Figure \ref{fig:WSDDA}.  
\begin{figure*}
\centering
\includegraphics[width=0.7 \textwidth]{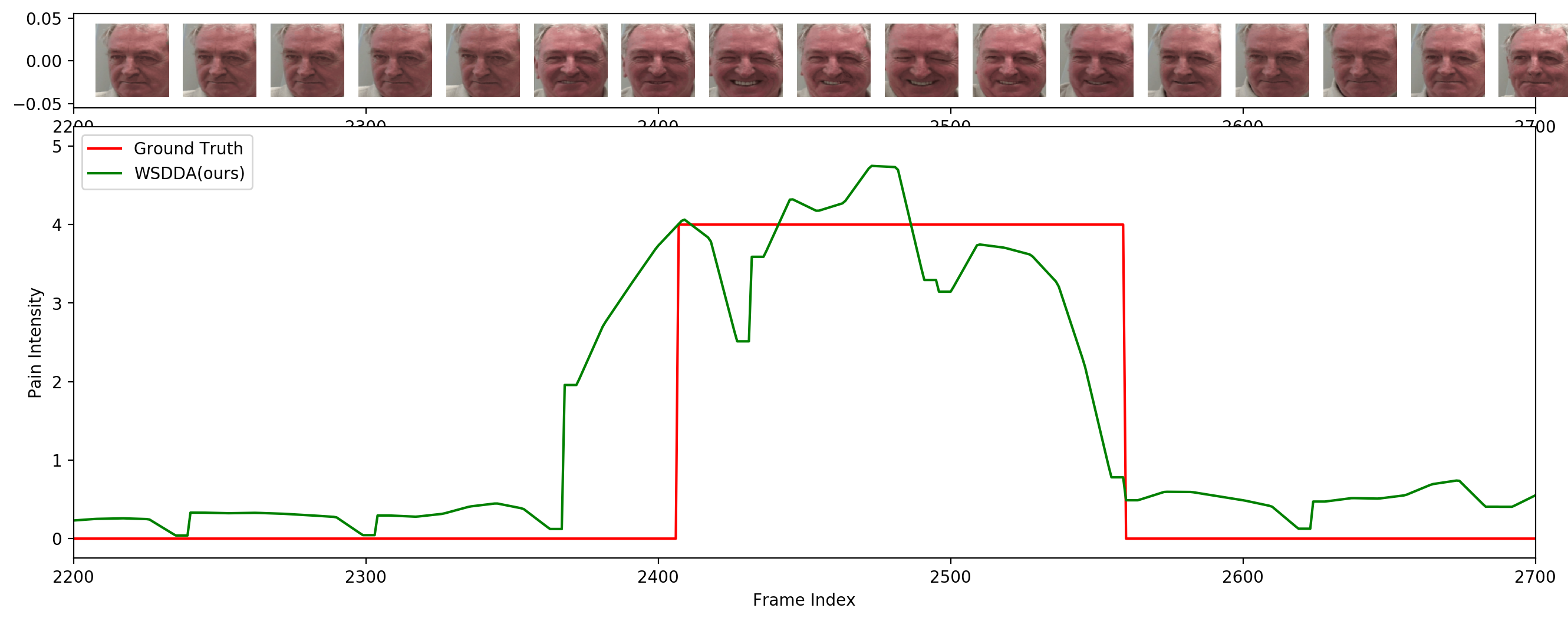}
\includegraphics[width=0.7 \textwidth]{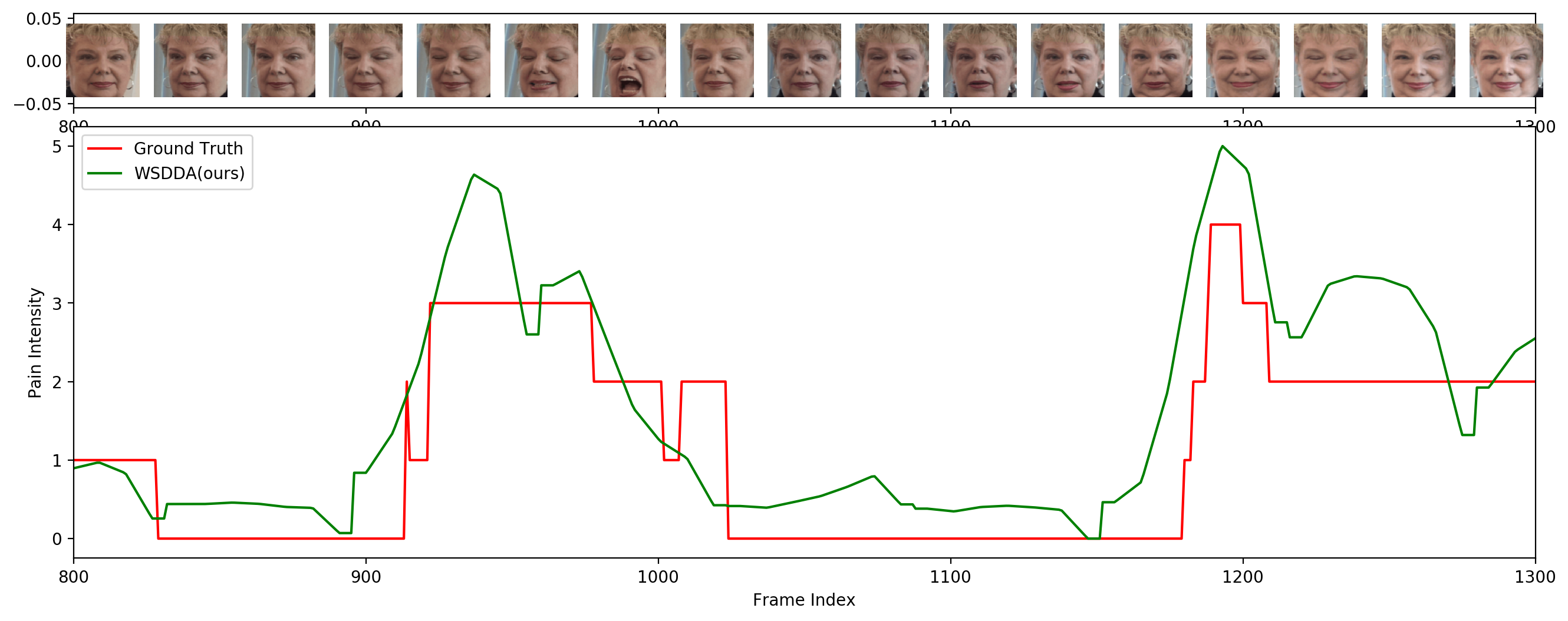}
\caption{\label{fig:painexp} \textbf{Visualization of pain localization on two different subjects. From top to bottom: scenario where ground truth (GT) shows no pain, but our deep WSDA approach correctly localizes pain. Scenario with multiple peaks of expressions}}
\label{pain loc vis}
\end{figure*}

\section{Results and Discussion}

\subsection{Experimental Setup:}
The proposed approach is evaluated on UNBC-McMaster data-set \cite{UNBC}, which has been widely used for pain estimation in the context of multiple instance learning. The data contains 200 videos of facial expressions captured from 25 individuals performing a series of active range of motion tests resulting in 47,398 frames of size 320x240. Each video sequence is annotated using PSPI score at frame-level on a range of 16 discrete pain intensity levels (0-15). In our experiments, we followed the same experimental protocol as that of \cite{midorf} in order to have fair comparison with the state-of-the-art results. Due to the high imbalance between low and high intensity levels, we followed the widely adapted quantization strategy i.e., the pain levels are quantized to 5 ordinal levels as: 0(0), 1(1), 2(2), 3(3), 4-5(4), 6-15(5). We have used Leave-One-Subject-Out (LOSO) cross-validation, where 15 subjects have been used for training, 9 subjects for validation and 1 for testing in each cycle.      

In our experiments, we have used I3D model, where inception v-1 architecture is the base model, which is inflated from 2D pre-trained model on ImageNet to 3D CNN for videos of facial expressions. Due to the availability of regression labels, RECOLA \cite{6553805} dataset is used as source data-set, where each video sequence has a duration of 5 minutes and is annotated with a regressed intensity value between -1 to +1 every 40 msec (same as frame rate of 25fps) i.e., all the frames are annotated. 
The video sequences of UNBC and RECOLA datasets are converted to sub-sequences of 64 frames with a stride of 8 to generate more number of samples for the learning framework, resulting in 10496 sub-sequences for RECOLA and 2890 sub-sequences for UNBC data-set. In order to incorporate the framework of weakly supervised learning, only coarse labels of the sub-sequences of UNBC data-set is considered by dropping some of the frame labels. The faces are detected, normalized and cropped using MTCNN \cite{MTCNN2016} and resized to 224 x 224. For training the model, we use Stochastic Gradient Descent (SGD) with a momentum of 0.9, weight decay of 1e5, and batch size of 8. The initial learning rate is set to 0.001 and annealed according to a schedule pre-determined on the cross validation set for every 5 epochs after 20 epochs. The maximum number of epochs is 50. An early stopping strategy is used to avoid over-fitting. The performance of the proposed approach is measured in terms of Pearson Correlation Coefficient (PCC), Intra class correlation (ICC(3,1)) and Mean-Average-Error (MAE). 

\subsection{Results with Baseline Training Models:}

The robustness of the proposed approach is validated by conducting a series of experiments with various baseline models, where I3D models are trained on data ranging from using only source data with full labels to the entire data-set of source and target domains with full labels as shown in Table \ref{results of various scenarios}. To reduce computational complexity, we have conducted a series of independent experiments by one of the target subjects for testing and rest of the subjects for training and validation. First, we train a model using only the source data with full labels without target domain, which is latter validated on test data of target domain. We can observe that the model exhibits poor performance due to the domain differences between train data and test data. Second, we train the model using only target data with full labels without source data and test the model on test data of target domain, which shows significant boost in performance as both train and test data are from same domain. Next, we evaluate the scenario where we train with both source and target domain with full labels, which further improves the performance of the system as training data spans wide range of variation due to domain differences of both data sets. 

\begin{table}
\hspace{-8mm}
\scriptsize
\renewcommand{\arraystretch}{1.4}
    \centering
\begin{tabular}{|l||c|c|c|c|c|c|c|c|c|c|} 
	\hline
	 \textbf{Training Scenario}  & \textbf{PCC} $\uparrow$ & \textbf{MAE} $\downarrow$  \\
	\hline 	\hline
    Supervised (source data only) & 0.295 & 1.630  \\
	\hline
	Supervised (target data only) & 0.447 &0.804 \\
	\hline
	Supervised (source $\cup$ target)  & 0.612 & 0.543 \\
	\hline 	\hline
	Unsupervised DA     & 0.413     & 0.874\\
	\hline
	WSDA (ours)         & 0.676     & 0.774\\
	\hline
	 Supervised DA      & 0.812     & 0.454 \\
	\hline
\end{tabular}
  \caption{ \textbf{PCC and MAE performance of I3D model trained under different scenarios.}}
    \label{results of various scenarios}
\end{table}

We conducted another series of experiments with domain adaptation, where the training data is provided with various levels of supervision for the target data but full labels of source data. In order to evaluate the performance of weakly supervised domain adaptation (WSDA), we have considered the scenarios of two extremes of the weak supervision i.e., full labels of target data (supervised) to no labels of target data (unsupervised), which acts as upper bound and lower bound respectively. 
By increasing the supervision of target domain, we can observe that the performance of the approach improves from unsupervised DA to supervised DA. By comparing the results of weak supervision with that of full supervision, it was found that the performance of the weakly supervised DA was found promising though target data is provided with weak coarse labels due to the fact that domain adaptation leverages the wide range of variation of both domains by minimizing the domain differences.

We have also further demonstrated the robustness of our approach by visualizing the predictions of pain intensity levels for some of the subjects as shown in Fig \ref{pain loc vis}. In the top plot of Fig \ref{pain loc vis}, we can observe that the proposed approach is able to recognize the onset of the facial expression though ground truth label fails to capture the onset of the expression and thereby efficiently localizes pain better than ground truth labels. Moreover, it proves that the proposed approach is able to relatively capture sudden transition in expressions though ground truth labels fails to do so. In the bottom plot of Fig \ref{pain loc vis}, multiple peaks of pain-expression events are presented, where our proposed approach efficiently follows the ground truth labels still retaining the gradual change in the onset and offset of expression. This shows that the proposed approach is able to capture multiple peaks of expressions even though they occur within short interval.

\begin{table*}
\scriptsize
\renewcommand{\arraystretch}{1.4}
    \centering
\begin{tabular}{|c|c|c|c|c|c|c|c|c|c|c|} 
	\hline
	 \textbf{Method}  & \textbf{Type of Supervision} & \multicolumn{3}{|c|}{\textbf{Frame-level}} & \multicolumn{3}{|c|}{\textbf{Sequence-level}}  \\ \cline{3-8}
	& & \textbf{PCC} $\uparrow$ & \textbf{MAE} $\downarrow$ & \textbf{ICC} $\uparrow$ & \textbf{PCC} $\uparrow$ & \textbf{MAE} $\downarrow$ & \textbf{ICC} $\uparrow$\\
	 \hline
	\hline
    MIR \cite{Aug MIR}  & weak & 0.350 &  0.840 & 0.240 & 0.63 & 0.940 & 0.630  \\
	\hline
	 MILBOOST \cite{sikka} & weak & 0.280 & 1.770 & 0.110 & 0.380 & 1.700 & 0.380\\
	\hline
	MI-DORF \cite{midorf}  & weak & 0.400 &  0.190 & 0.460 & 0.670 & 0.800 & 0.660\\
	\hline
	 Deep WSDA (ours) & weak & \textbf{0.630}  & \textbf{0.714} & \textbf{0.567} & \textbf{0.828} & \textbf{0.647} & \textbf{0.762}\\ 
	\hline \hline
	BORMIR \cite{zhangbilateral} & semi & 0.605  & 0.821 & 0.531 & - & - & - \\
	\hline \hline
	LSTM \cite{7849133} &  full &0.780  & 0.500 & - & - & - & -\\
	\hline
	SCN \cite{Tavakolian2018DeepSR} & full  &0.920  & 0.320 (MSE) & 0.750 & - & - & - \\
	\hline
\end{tabular}
  \caption{ \textbf{PCC, MAE and ICC performance of proposed and state-of-art methods trained under different scenarios.}}
    \label{Comparison with state-of-the-art}
\end{table*}

\begin{figure}
\centering
\includegraphics[width=0.9\linewidth]{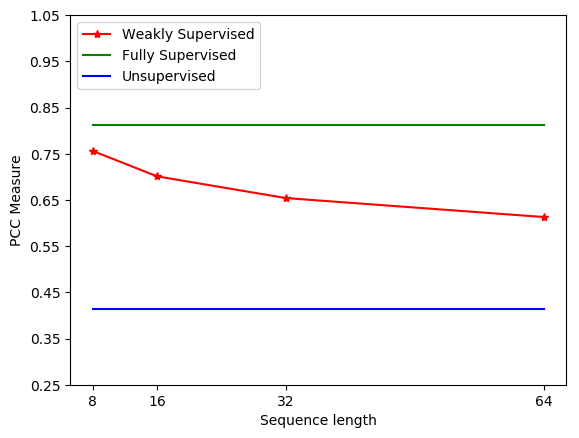}
\caption{\textbf{PCC accuracy of I3D model trained with deep WSDA levels with decreasing level of weak supervision on target videos.}}
\label{fig:painloc}
\end{figure}

We have also analyzed the performance of our approach for pain localization of videos, ie., instance level prediction for varying levels of supervision by changing the frequency of annotations for sequences (bags). By considering the full labels of source domain, labels of target domain are gradually reduced by decreasing the frequency of annotations i.e., labels are provided for longer duration of sequences. As we gradually reduce the amount of labels of the target domain, we can observe that the performance of our approach gradually drops, which is depicted in Fig \ref{fig:painloc}. Specifically, we have conducted experiments for sequence lengths of 8,16,32 and 64. However, our approach shows promising performance even with minimal annotation, which is attributed to the domain adaptation as we are leveraging source data to adapt to target domain using adversarial domain adaptation. Therefore, i3D model in conjunction with domain adaptation was found to achieve superior results even with minimal supervision of target data. 

In addition to the frequency of annotations, there are several factors that can inﬂuence the performance of our proposed system. One of the major factors is the amount of domain shift between source and target domains, which plays a crucial role in learning discriminative domain invariant feature embeddings, and improving the performance of the model in target domain. Another major factor is the proportion of the size of source and target domain data-sets, which controls the trade-off between predictive learning and learning domain invariant features. Finally, some of the prevalent factors such as frequency of relevant pain expressions, subjective variations in the data-set, etc, also impacts the performance of the system.

\subsection{Comparison with State-of-Art Methods:}

Traditional machine learning approaches have been explored in most of the state-of-the-art approaches for weakly supervised learning due to the problem of limited data with limited annotations. However, we have used deep learning based I3D model along with source data to compensate the problem of limited data using domain adaptation. We have also provided visualization of some of our results for pain localization i.e., frame-level prediction for two subjects (see Figure \ref{pain loc vis}). We can observe that the proposed approach correctly localizes pain even though it was not provided in ground truth.

Given the close relation, we compare our work with \cite{midorf}, which uses graph based models to capture the temporal relationship of the frames and ordinal relationship of labels. Due to the limited work of regression based methods in the context of multiple instance learning, we have also compared our approach with that of \cite{sikka}, which was proposed for pain classification. In order to compare with classification based methods which predicts only the presence or absence of pain, the output probability for the presence of pain is considered as indicator of pain intensity levels. That is higher probabilities in pain prediction denotes higher pain intensity level, where the output probability for occurrence of pain is normalized between 0 and 5, which is also discussed in \cite{midorf}. Among the approaches based on multiple instance learning, we can observe that \cite{midorf} outperforms \cite{sikka} and \cite{Aug MIR} due to the fact that classification based methods fails to accurately predict the pain intensities, whereas \cite{midorf} proposed a regression framework for estimating pain intensities. On the contrary, the proposed approach significantly outperforms these methods in terms of all the performance metrics by leveraging the potential of deep models and domain adaptation.
Among all the performance metrics, PCC measures the inter correlation between the ground truth and predicted outputs and thereby exhibits best performance compared to ICC and MAE. ICC is mostly used to measure the degree of correlation and agreement between the measurements, whereas MAE is widely used for predictive regression tasks. Since ICC is more reliable than PCC for sequence-level estimation as it efficiently captures the intraclass correlation, the proposed approach exhibits higher performance of ICC for sequence-level estimation compared to frame-level estimation.

In addition to MIL based approaches, we have also compared our results with the state-of-the-art methods for fully supervised and semi-supervised scenarios. In the case of partial annotations, though \cite{zhangbilateral} uses deep models for estimation of facial action unit intensity estimation, it requires the location of peak and valley frames for training. However, our approach performs at par with \cite{zhangbilateral} without the need to locate peak and valley frames. By comparing with the the state-of-art approach results as shown in Table \ref{Comparison with state-of-the-art}, the performance of the proposed approach is quite promising as it leverages the deep I3D models along with domain adaptation. 


\section{Conclusion and Future Work}
In this paper, we have proposed a deep WSDA technique for training 3D-CNNs for spatio-temporal pain intensity estimation from weakly-labeled videos. We have conducted extensive set of experiments with various baseline models using various combinations of source and target data-set and evaluated the performance of our proposed approach over baseline models. We have also compared our approach to the state-of-the-art weakly supervised as well as fully supervised methods and shown that the proposed approach is quite promising in handling the problem of data with limited annotations over state-of-the-art approaches. 

Finally, we have analyzed the performance of our approach under varying levels of supervision for the target data. From our investigations, it was found that temporal modeling seems to be promising in generating efficient predictive models for facial expression analysis in videos and adversarial domain adaptation efficiently handles the problem of data with limited annotations. Since DL models are still at rudimentary level for weakly supervised scenarios in the context of multiple instance learning, there is a lot of room for improving the performance of the system by overcoming the problem of limited annotations still leveraging 3D CNN models. 

\end{document}